\documentclass[10pt, a4paper]{article}
\usepackage{amssymb}
\usepackage{lrec}
\usepackage{multibib}
\usepackage{graphicx}
\usepackage{tabularx}
\usepackage{soul}
\usepackage{amssymb}

\usepackage{epstopdf}
\usepackage[latin1]{inputenc}
\usepackage{multirow}
\usepackage{hyperref}
\usepackage{xstring}
\usepackage{CJK}

\title{The First Evaluation of Chinese Human-Computer Dialogue Technology}

\name{Wei-Nan Zhang$^{\dagger}$, Zhigang Chen$^{\ddagger}$, Wanxiang Che$^{\dagger}$, Guoping Hu$^{\ddagger}$, Ting Liu$^{\dagger}$}

\address{$\dagger$Research Center for Social Computing and Information Retrieval, Harbin Institute of Technology, Harbin, China \\
         $\ddagger$Joint Laboratory of HIT and iFLYTEK, iFLYTEK Research, Hefei, China \\
          \{wnzhang, car, tliu\}@ir.hit.edu.cn, \{zgchen, gphu\}@iflytek.com\\}

\abstract{
In this paper, we introduce the first evaluation of Chinese human-computer dialogue technology.
We detail the evaluation scheme, tasks, metrics and how to collect and annotate the data for training, developing and test.
The evaluation includes two tasks, namely \textbf{user intent classification} and \textbf{online testing of task-oriented dialogue}.
To consider the different sources of the data for training and developing, the first task can also be divided into two sub tasks.
Both the two tasks are coming from the real problems when using the applications developed by industry.
The evaluation data is provided by the iFLYTEK Corporation.
Meanwhile, in this paper, we publish the evaluation results to present the current performance of the participants in the two tasks of Chinese human-computer dialogue technology. Moreover, we analyze the existing problems of human-computer dialogue as well as the evaluation scheme itself.
\\
\newline  }

\begin{document}
\begin{CJK*}{UTF8}{gbsn}

\maketitleabstract

\section{Introduction}

Recently, human-computer dialogue has been emerged as a hot topic, which has attracted the attention of both academia and industry.
In research, the natural language understanding (NLU), dialogue management (DM) and natural language generation (NLG) have been promoted by the technologies of big data and deep learning~\cite{7,8,9,10,11,12,13,14,15}.
Following the development of machine reading comprehension~\cite{16,17,18,19,20,aoa}, the NLU technology has made great progress.
The development of DM technology is from rule-based approach and supervised learning based approach to reinforcement learning based approach~\cite{24}.
The NLG technology is through pattern-based approach, sentence planning approach and end-to-end deep learning approach~\cite{21,22,23}.
In application, there are massive products that are based on the technology of human-computer dialogue, such as Apple Siri\footnote{https://www.apple.com/ios/siri/}, Amazon Echo\footnote{https://en.wikipedia.org/wiki/Amazon\_Echo}, Microsoft Cortana\footnote{https://www.microsoft.com/en-us/windows/cortana}, Facebook Messenger\footnote{https://en.wikipedia.org/wiki/Facebook\_Messenger} and Google Allo\footnote{https://allo.google.com/} etc.

\begin{figure}[!h]
\begin{center}
\includegraphics[width=0.5\textwidth]{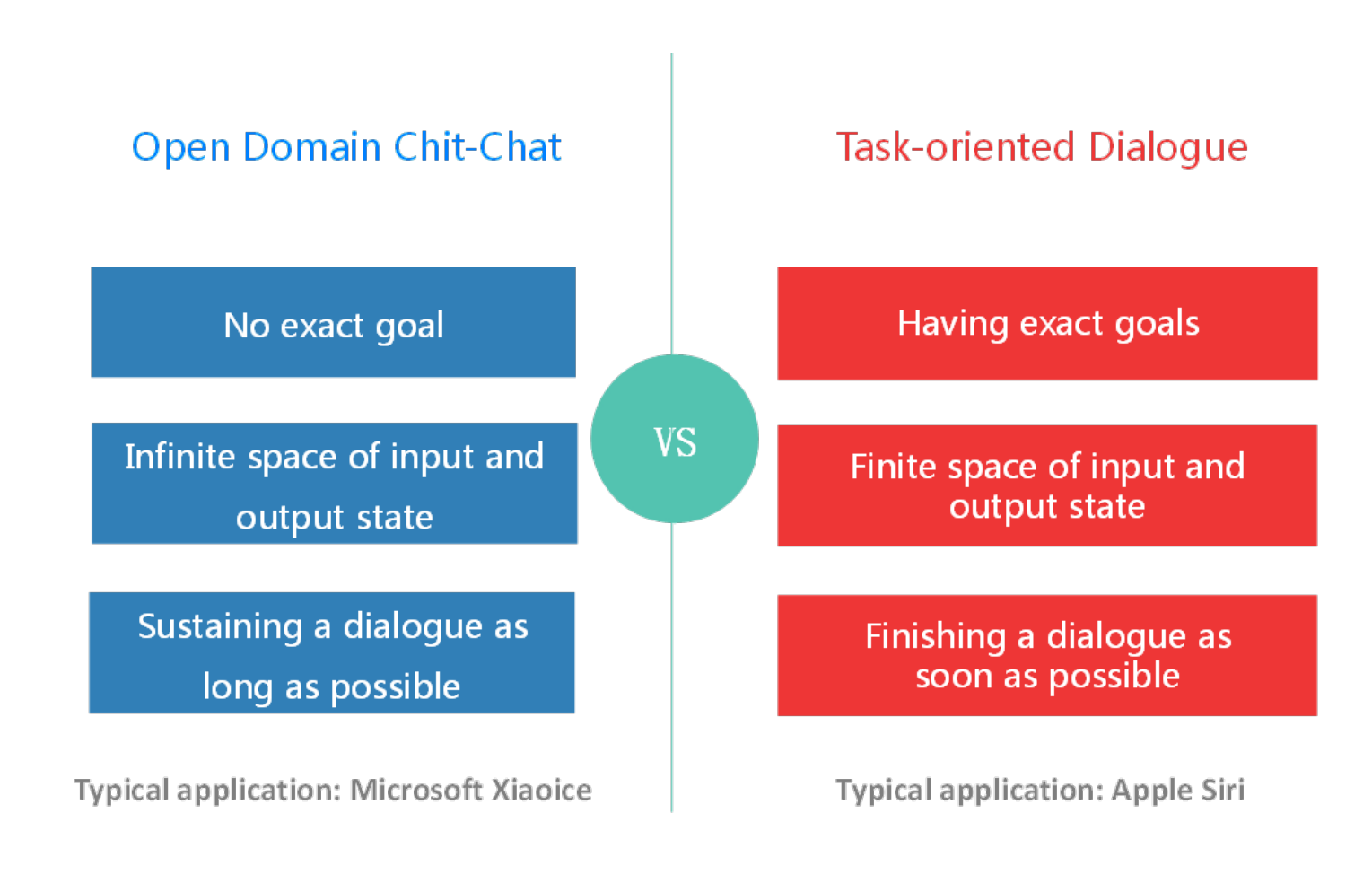}
\caption{A brief comparison of the open domain chit-chat system and the task-oriented dialogue system.}
\label{fig.1}
\end{center}
\end{figure}

Although the blooming of human-computer dialogue technology in both academia and industry, how to evaluate a dialogue system, especially an open domain chit-chat system, is still an open question.
Figure~\ref{fig.1} presents a brief comparison of the open domain chit-chat system and the task-oriented dialogue system.

\begin{table*}[!t]
\begin{center}
\begin{tabular}{|l|l|}
\hline
Input message & Intent category \\
\hline
你好啊，很高兴见到你！& 闲聊类 \\
Hello, nice to meet you! & Chit-chat \\
\hline
我想订一张去北京的机票 & 任务型（订机票） \\
I want to book an air ticket to Beijing. & Task-oriented dialogue (Booking air tickets) \\
\hline
我想找一家五道口附近便宜干净的快捷酒店 & 任务型（订酒店） \\
I want to book a neat and low-priced inn near Wudaokou. & Task-oriented dialogue (Booking hotels) \\
\hline
\end{tabular}
\caption{An example of user intent with category information. \label{tab.1}}
\end{center}
\end{table*}

From Figure~\ref{fig.1}, we can see that it is quite different between the open domain chit-chat system and the task-oriented dialogue system.
For the open domain chit-chat system, as it has no exact goal in a conversation, given an input message, the responses can be various.
For example, for the input message \emph{``How is it going today?''}, the responses can be \emph{``I'm fine!''}, \emph{``Not bad.''}, \emph{``I feel so depressed!''}, \emph{``What a bad day!''}, etc.
There may be infinite number of responses for an open domain messages.
Hence, it is difficult to construct a gold standard (usually a reference set) to evaluate a response which is generated by an open domain chit-chat system.
For the task-oriented system, although there are some objective evaluation metrics, such as the number of turns in a dialogue\footnote{The number of utterances in a task-completed dialogue. Given a same task, the less number of turns, the better of the dialogue system.}, the ratio of task completion\footnote{The number of completed tasks divided by the total number of tasks.}, etc.,
there is no gold standard for automatically evaluating two (or more) dialogue systems when considering the satisfaction of the human and the fluency of the generated dialogue.

To promote the development of the evaluation technology for dialogue systems, especially considering the language characteristics of Chinese, we organize the first evaluation of Chinese human-computer dialogue technology.
In this paper, we will present the evaluation scheme and the released corpus in detail.

The rest of this paper is as follows. In Section 2, we will briefly introduce the first evaluation of Chinese human-computer dialogue technology, which includes the descriptions and the evaluation metrics of the two tasks.
We then present the evaluation data and final results in Section 3 and 4 respectively, following the conclusion and acknowledgements in the last two sections.

\section{The First Evaluation of Chinese Human-Computer Dialogue Technology}

The First Evaluation of Chinese Human-Computer Dialogue Technology includes two tasks, namely \textbf{user intent classification} and \textbf{online testing of task-oriented dialogue}.
\subsection{Task 1: User Intent Classification}\label{task1}

In using of human-computer dialogue based applications, human may have various intent, for example, chit-chatting, asking questions, booking air tickets, inquiring weather, etc. Therefore, after receiving an input message (text or ASR result) from a user, the first step is to classify the user intent into a specific domain for further processing.
Table~\ref{tab.1} shows an example of user intent with category information.

In task 1, there are two top categories, namely, chit-chat and task-oriented dialogue.
The task-oriented dialogue also includes 30 sub categories.
In this evaluation, we only consider to classify the user intent in single utterance.

It is worth noting that besides the released data for training and developing, we also allow to collect external data for training and developing.
To considering that, the task 1 is indeed includes two sub tasks.
One is a \textbf{closed evaluation}, in which only the released data can be used for training and developing.
The other is an \textbf{open evaluation} that allow to collect external data for training and developing.
For task 1, we use F1-score as evaluation metric.

\subsection{Task 2: Online Testing of Task-oriented Dialogue}

For the task-oriented dialogue systems, the best way for evaluation is to use the online human-computer dialogue.
After finishing an online human-computer dialogue with a dialogue system, the human then manually evaluate the system by using the metrics of user satisfaction degree, dialogue fluency, etc.
Therefore, in the task 2, we use an online testing of task-oriented dialogue for dialogue systems.
For a human tester, we will give a complete intent with an initial sentence, which is used to start the online human-computer dialogue.
Table~\ref{tab.2} shows an example of the task-oriented human-computer dialogue. Here ``U'' and ``R'' denote user and robot respectively.
The complete intent is as following: \\ \\
\emph{``查询明天从哈尔滨到北京的晚间软卧火车票，上下铺均可。 \\ Inquire the soft berth ticket at tomorrow evening, from Harbin to Beijing, either upper or lower berth is okay.''}\\

\begin{table}[!t]
\begin{center}
\begin{tabular}{|l|l|}
\hline
\multirow{2}{*}{U} & 有到北京的软卧火车票吗？ \\
& Are there any soft berth tickets to Beijing? \\
\hline
\multirow{2}{*}{R} & 请问您要从哪里出发？ \\
& Where do you leave from? \\
\hline
\multirow{2}{*}{U} & 哈尔滨 \\
& Harbin. \\
\hline
\multirow{2}{*}{R} & 您要订哪天的呢？ \\
& When will you leave? \\
\hline
\multirow{2}{*}{U} & 明天吧 \\
& Tomorrow. \\
\hline
\multirow{2}{*}{R} & 发车时间呢？ \\
& What about the departure time? \\
\hline
\multirow{2}{*}{U} & 晚上吧，正好睡觉。\\
& Evening is okay, just sleep. \\
\hline
\multirow{2}{*}{R} & 上铺还是下铺？ \\
& Do you want an upper or lower berth? \\
\hline
\multirow{2}{*}{U} & 都行。 \\
& Either is okay. \\
\hline
\multirow{2}{*}{R} & 以下是为您查询到的车票信息。 \\
& The tickets information is as following. \\
\hline
\multirow{2}{*}{U} & 谢谢，再见。 \\
& Thanks, see you. \\
\hline
\multirow{2}{*}{R} & 不客气，再见。 \\
& You're welcome. Goodbye. \\
\hline
\end{tabular}
\caption{An example of the task-oriented human-computer dialogue. \label{tab.2}}
\end{center}
\end{table}

In task 2, there are three categories.
They are ``air tickets'', ``train tickets'' and ``hotel''.
Correspondingly, there are three type of tasks.
All the tasks are in the scope of the three categories.
However, a complete user intent may include more than one task.
For example, a user may first inquiring the air tickets.
However, due to the high price, the user decide to buy a train tickets.
Furthermore, the user may also need to book a hotel room at the destination.

\begin{table*}[!t]
\begin{center}
\begin{tabular}{|l|c|c|c|c|c|c|c|c|c|c|c|}
\hline
 & app & bus & calc & \textbf{chat} & cinemas & contacts & cookbook & datetime & email & epg & flight \\
\hline
Train & 36 & 24 & 24 & 456 & 24 & 30 & 269 & 18 & 24 & 107 & 62 \\
Dev & 18 & 8 & 8 & 114 & 10 & 10 & 88 & 6 & 8 & 36 & 21 \\
Test & 18 & 8 & 8 & 50 & 8 & 10 & 90 & 6 & 8 & 36 & 21 \\
Sum & 72 & 40 & 40 & 662 & 42 & 50 & 447 & 30 & 40 & 179 & 104 \\
\hline
& health & lottery & map & match & message & music & news & novel & poetry & radio & riddle \\
\hline
Train & 55 & 24 & 68 & 24 & 63 & 66 & 58 & 24 & 402 & 24 & 34 \\
Dev & 19 & 8 & 23 & 8 & 21 & 22 & 19 & 8 & 34 & 8 & 11 \\
Test & 18 & 8 & 23 & 8 & 21 & 22 & 19 & 8 & 34 & 8 & 11 \\
Sum & 92 & 40 & 114 & 40 & 105 & 110 & 96 & 40 & 470 & 40 & 56 \\
\hline
& schedule & stock & telephone & train & translation & tvchannel & video & weather & website & \multicolumn{2}{c|}{Total \#} \\
\hline
Train & 29 & 71 & 63 & 70 & 61 & 71 & 182 & 66 & 54  & \multicolumn{2}{c|}{2,583} \\
Dev & 9 & 24 & 21 & 23 & 21 & 23 & 60 & 22 & 18  & \multicolumn{2}{c|}{729} \\
Test & 10 & 24 & 21 & 23 & 20 & 24 & 61 & 22 & 18  & \multicolumn{2}{c|}{688} \\
Sum & 48 & 119 & 105 & 116 & 102 & 118 & 303 & 110 & 90  & \multicolumn{2}{c|}{4,000} \\
\hline
\end{tabular}
\caption{The statistics of the released data for task 1. \label{tab.3}}
\end{center}
\end{table*}

\begin{table*}[!t]
\begin{center}
\begin{tabular}{|c|c|c|}
\hline
Ranking & Participant & F1 score \\
\hline
\multirow{2}{*}{1} & 华南农业大学口语对话系统研究室 & \multirow{2}{*}{0.9391} \\
& Spoken Dialogue System Lab, South China Agricultural University &  \\
\hline
\multirow{2}{*}{2} & 义语智能科技（上海）有限公司 & \multirow{2}{*}{0.9288} \\
& DeepBrain Corporation &  \\
\hline
\multirow{2}{*}{3} & 山西大学计算机与信息技术学院 & \multirow{2}{*}{0.9089} \\
& School of Computer \& Information Technology, Shanxi University &  \\
\hline
\multirow{2}{*}{4} & 北京邮电大学智能科学与技术中心 & \multirow{2}{*}{0.9082} \\
& Intelligent science and technology center, Beijing University of Posts and Telecommunications &  \\
\hline
\multirow{2}{*}{5} & 哈尔滨工业大学(深圳) & \multirow{2}{*}{0.9028} \\
& Harbin Institute of Technology (Shenzhen) &  \\
\hline
\end{tabular}
\caption{Top 5 results of the closed test of the task 1. \label{tab.4}}
\end{center}
\end{table*}

\begin{table*}[!t]
\begin{center}
\begin{tabular}{|c|c|c|}
\hline
Ranking & Participant & F1 score \\
\hline
\multirow{2}{*}{1} & 华南农业大学口语对话系统研究室 & \multirow{2}{*}{0.9414} \\
& Spoken Dialogue System Lab, South China Agricultural University &  \\
\hline
\multirow{2}{*}{2} & 义语智能科技（上海）有限公司 & \multirow{2}{*}{0.9288} \\
& DeepBrain Corporation &  \\
\hline
\multirow{2}{*}{3} & 中国科学院自动化研究所-出门问问语言智能与人机交互联合实验室 & \multirow{2}{*}{0.9258} \\
& Institute of automation, Chinese Academy of Sciences &  \\
\hline
\multirow{2}{*}{4} & 广东外语外贸大学 & \multirow{2}{*}{0.9255} \\
& Guangdong University of Foreign Studies &  \\
\hline
\multirow{2}{*}{5} & 山西大学计算机与信息技术学院 & \multirow{2}{*}{0.9123} \\
& School of Computer \& Information Technology, Shanxi University &  \\
\hline
\end{tabular}
\caption{Top 5 results of the open test of the task 1. \label{tab.5}}
\end{center}
\end{table*}

\begin{table*}[!t]
\begin{center}
\begin{tabular}{|c|c|c|c|c|c|c|c|}
\hline
Ranking & Participant  & Ratio & Satisfaction & Fluency & Turns & Guide \\
\hline
\multirow{2}{*}{1} & 深思考人工智能机器人科技（北京）有限公司 & \multirow{2}{*}{0.3175} & \multirow{2}{*}{64.53} & \multirow{2}{*}{0} & \multirow{2}{*}{-1} & \multirow{2}{*}{2} \\
& iDeepMind Artificial Intelligence & & & & &  \\
\hline
\multirow{2}{*}{2} & 上海葡萄纬度科技有限公司 & \multirow{2}{*}{0.1905} & \multirow{2}{*}{72.28} & \multirow{2}{*}{-1} & \multirow{2}{*}{1} & \multirow{2}{*}{3} \\
& Shanghai Putao Technology Co., Ltd. & & & & &  \\
\hline
\multirow{3}{*}{3} & 北京邮电大学信息与通信工程学院 & \multirow{3}{*}{0.1905} & \multirow{3}{*}{78.72} & \multirow{3}{*}{0} & \multirow{3}{*}{1} & \multirow{3}{*}{3} \\
& School of Information and Communication Engineering & & & & &  \\
& Beijing University of Posts and Telecommunications & & & & &  \\
\hline
\multirow{3}{*}{4} & 中国科学院自动化研究所 & \multirow{3}{*}{0.1111} & \multirow{3}{*}{71.39} & \multirow{3}{*}{-2} & \multirow{3}{*}{-1} & \multirow{3}{*}{3} \\
& 出门问问语言智能与人机交互联合实验室  & & & & &   \\
& Institute of automation, Chinese Academy of Sciences  & & & & &   \\
\hline
\end{tabular}
\caption{The results of the task 2. Ratio, Satisfaction, Fluency, Turns and Guide indicate the task completion ratio, user satisfaction degree, response fluency, number of dialogue turns and guidance ability for out of scope input respectively. \label{tab.6}}
\end{center}
\end{table*}

We use manual evaluation for task 2.
For each system and each complete user intent, the initial sentence, which is used to start the dialogue, is the same.
The tester then begin to converse to each system.
A dialogue is finished if the system successfully returns the information which the user inquires or the number of dialogue turns is larger than 30 for a single task.
For building the dialogue systems of participants, we release an example set of complete user intent and three data files of flight, train and hotel in JSON format.
There are five evaluation metrics for task 2 as following.
\begin{itemize}
  \item \textbf{Task completion ratio}: The number of completed tasks divided by the number of total tasks.
  \item \textbf{User satisfaction degree}: There are five scores -2, -1, 0, 1, 2, which denote very dissatisfied, dissatisfied, neutral, satisfied and very satisfied, respectively.
  \item \textbf{Response fluency}: There are three scores -1, 0, 1, which indicate nonfluency, neutral, fluency.
  \item \textbf{Number of dialogue turns}: The number of utterances in a task-completed dialogue.
  \item \textbf{Guidance ability for out of scope input}\footnote{During the dialogue, the testers may inquire the information that is not existing in the data files. These inquiry inputs are called out of scope inputs.}: There are two scores 0, 1, which represent able to guide or unable to guide.
\end{itemize}

For the number of dialogue turns, we have a penalty rule that for a dialogue task, if the system cannot return the result (or accomplish the task) in 30 turns, the dialogue task is end by force.
Meanwhile, if a system cannot accomplish a task in less than 30 dialogue turns, the number of dialogue turns is set to 30.

\section{Evaluation Data}

In the evaluation, all the data for training, developing and test is provided by the iFLYTEK Corporation\footnote{http://www.iflytek.com/en/index.html}.

For task 1, as the descriptions in Section~\ref{task1}, the two top categories are chit-chat (\textbf{chat} in Table~\ref{tab.3}) and task-oriented dialogue.
Meanwhile, the task-oriented dialogue also includes 30 sub categories.
Actually, the task 1 is a 31 categories classification task.
In task 1, besides the data we released for training and developing, we also allow the participants to extend the training and developing corpus.
Hence, there are two sub tasks for the task 1.
One is closed test, which means the participants can only use the released data for training and developing.
The other is open test, which allows the participants to explore external corpus for training and developing.
Note that there is a same test set for both the closed test and the open test.

For task 2, we release 11 examples of the complete user intent and 3 data file, which includes about one month of flight, hotel and train information, for participants to build their dialogue systems.
The current date for online test is set to April 18, 2017.
If the tester says ``today'', the systems developed by the participants should understand that he/she indicates the date of April 18, 2017.

\section{Evaluation Results}

There are 74 participants who are signing up the evaluation.
The final number of participants is 28 and the number of submitted systems is 43.
Table~\ref{tab.4} and~\ref{tab.5} show the evaluation results of the closed test and open test of the task 1 respectively.
Due to the space limitation, we only present the top 5 results of task 1.
We will add the complete lists of the evaluation results in the version of full paper.

Note that for task 2, there are 7 submitted systems.
However, only 4 systems can provide correct results or be connected in a right way at the test phase.
Therefore, Table~\ref{tab.6} shows the complete results of the task 2.

\section{Conclusion}

In this paper, we introduce the first evaluation of Chinese human-computer dialogue technology.
In detail, we first present the two tasks of the evaluation as well as the evaluation metrics.
We then describe the released data for evaluation.
Finally, we also show the evaluation results of the two tasks.
As the evaluation data is provided by the iFLYTEK Corporation from their real online applications, we believe that the released data will further promote the research of human-computer dialogue and fill the blank of the data on the two tasks.

\section*{Acknowledgements}

We would like to thank the Social Media Processing (SMP) committee of Chinese Information Processing Society of China.
We thank all the participants of the first evaluation of Chinese human-computer dialogue technology.
We also thank the testers from the voice resource department of the iFLYTEK Corporation for their effort to the online real-time human-computer dialogue test and offline dialogue evaluation.
We thank Lingzhi Li, Yangzi Zhang, Jiaqi Zhu and Xiaoming Shi from the research center for social computing and information retrieval for their support on the data annotation, establishing the system testing environment and the communication to the participants and help connect their systems to the testing environment.

This paper is supported by National Key Basic Research Program of China (No. 2014CB340503) and NSFC (No. 61502120), the Fundamental Research Funds for the Central Universities(30620170037).

\section*{Reference}
\label{main:ref}
\bibliographystyle{lrec}
\bibliography{xample}

\begin{thebibliography}{}

\bibitem[\protect\citename{Cui \bgroup et al.\egroup }2016]{16}
Cui, Y., Liu, T., Chen, Z., Wang, S., and Hu, G.
\newblock (2016).
\newblock Consensus attention-based neural networks for chinese reading
  comprehension.
\newblock In {\em Proceedings of COLING 2016, the 26th International Conference
  on Computational Linguistics: Technical Papers}, pages 1777--1786, Osaka,
  Japan.

\bibitem[\protect\citename{Cui \bgroup et al.\egroup }2017]{aoa}
Cui, Y., Chen, Z., Wei, S., Wang, S., Liu, T., and Hu, G.
\newblock (2017).
\newblock Attention-over-attention neural networks for reading comprehension.
\newblock In {\em Proceedings of the 55th Annual Meeting of the Association for
  Computational Linguistics (Volume 1: Long Papers)}, pages 593--602.
  Association for Computational Linguistics.

\bibitem[\protect\citename{Hermann \bgroup et al.\egroup }2015]{17}
Hermann, K.~M., Kocisky, T., Grefenstette, E., Espeholt, L., Kay, W., Suleyman,
  M., and Blunsom, P.
\newblock (2015).
\newblock Teaching machines to read and comprehend.
\newblock In {\em Advances in Neural Information Processing Systems}, pages
  1693--1701.

\bibitem[\protect\citename{Kadlec \bgroup et al.\egroup }2016]{18}
Kadlec, R., Schmid, M., Bajgar, O., and Kleindienst, J.
\newblock (2016).
\newblock Text understanding with the attention sum reader network.
\newblock In {\em Proceedings of the 54th Annual Meeting of the Association for
  Computational Linguistics (Volume 1: Long Papers)}, pages 908--918, Berlin,
  Germany, August. Association for Computational Linguistics.

\bibitem[\protect\citename{Li \bgroup et al.\egroup }2015]{10}
Li, J., Galley, M., Brockett, C., Gao, J., and Dolan, B.
\newblock (2015).
\newblock A diversity-promoting objective function for neural conversation
  models.
\newblock {\em NAACL}.

\bibitem[\protect\citename{Li \bgroup et al.\egroup }2016]{12}
Li, J., Monroe, W., Ritter, A., and Dan, J.
\newblock (2016).
\newblock Deep reinforcement learning for dialogue generation.
\newblock {\em EMNLP}.

\bibitem[\protect\citename{Liu \bgroup et al.\egroup }2017]{19}
Liu, T., Cui, Y., Yin, Q., Zhang, W.-N., Wang, S., and Hu, G.
\newblock (2017).
\newblock Generating and exploiting large-scale pseudo training data for zero
  pronoun resolution.
\newblock In {\em Proceedings of the 55th Annual Meeting of the Association for
  Computational Linguistics (Volume 1: Long Papers)}, pages 102--111,
  Vancouver, Canada, July. Association for Computational Linguistics.

\bibitem[\protect\citename{Mairesse and Young}2014]{21}
Mairesse, F. and Young, S.
\newblock (2014).
\newblock Stochastic language generation in dialogue using factored language
  models.
\newblock {\em Computational Linguistics}, 40(4):763--799.

\bibitem[\protect\citename{Mairesse \bgroup et al.\egroup }2010]{22}
Mairesse, F., ~, M., Jurcicek, Ek, F., Keizer, S., Thomson, B., Yu, K., and
  Young, S.
\newblock (2010).
\newblock Phrase-based statistical language generation using graphical models
  and active learning.
\newblock In {\em ACL}, pages 1552--1561.

\bibitem[\protect\citename{Serban \bgroup et al.\egroup }2016a]{8}
Serban, I.~V., Sordoni, A., Bengio, Y., Courville, A., and Pineau, J.
\newblock (2016a).
\newblock Building end-to-end dialogue systems using generative hierarchical
  neural network models.
\newblock {\em Computer Science}.

\bibitem[\protect\citename{Serban \bgroup et al.\egroup }2016b]{9}
Serban, I.~V., Sordoni, A., Lowe, R., Charlin, L., Pineau, J., Courville, A.,
  and Bengio, Y.
\newblock (2016b).
\newblock A hierarchical latent variable encoder-decoder model for generating
  dialogues.

\bibitem[\protect\citename{Shang \bgroup et al.\egroup }2015]{7}
Shang, L., Lu, Z., and Li, H.
\newblock (2015).
\newblock Neural responding machine for short-text conversation.
\newblock In {\em ACL}, pages 1577--1586.

\bibitem[\protect\citename{Vinyals and Le}2015]{13}
Vinyals, O. and Le, Q.
\newblock (2015).
\newblock A neural conversational model.
\newblock {\em Computer Science}.

\bibitem[\protect\citename{Wang \bgroup et al.\egroup }2017]{20}
Wang, W., Yang, N., Wei, F., Chang, B., and Zhou, M.
\newblock (2017).
\newblock Gated self-matching networks for reading comprehension and question
  answering.
\newblock In {\em Proceedings of the 55th Annual Meeting of the Association for
  Computational Linguistics (Volume 1: Long Papers)}, volume~1, pages 189--198.

\bibitem[\protect\citename{Wen \bgroup et al.\egroup }2015a]{23}
Wen, T.~H., Gasic, M., Kim, D., Mrksic, N., Su, P.~H., Vandyke, D., and Young,
  S.
\newblock (2015a).
\newblock Stochastic language generation in dialogue using recurrent neural
  networks with convolutional sentence reranking.
\newblock {\em SIGDial}.

\bibitem[\protect\citename{Wen \bgroup et al.\egroup }2015b]{15}
Wen, T.~H., Gasic, M., Mrksic, N., Su, P.~H., Vandyke, D., and Young, S.
\newblock (2015b).
\newblock Semantically conditioned lstm-based natural language generation for
  spoken dialogue systems.
\newblock {\em EMNLP}.

\bibitem[\protect\citename{Wen \bgroup et al.\egroup }2016]{14}
Wen, T.-H., Ga\v{s}i\'{c}, M., Mrk\v{s}i\'{c}, N., Rojas-Barahona, L.~M., Su,
  P.-H., Vandyke, D., and Young, S.
\newblock (2016).
\newblock Multi-domain neural network language generation for spoken dialogue
  systems.
\newblock In {\em NAACL}, pages 120--129.

\bibitem[\protect\citename{Xing \bgroup et al.\egroup }2016]{11}
Xing, C., Wu, W., Wu, Y., Liu, J., Huang, Y., Zhou, M., and Ma, W.~Y.
\newblock (2016).
\newblock Topic augmented neural response generation with a joint attention
  mechanism.

\bibitem[\protect\citename{Young \bgroup et al.\egroup }2013]{24}
Young, S., Gasic, M., Thomson, B., and Williams, J.~D.
\newblock (2013).
\newblock Pomdp-based statistical spoken dialog systems: A review.
\newblock {\em Proceedings of the IEEE}, pages 1160--1179.

\end{thebibliography}

\end{CJK*}
\end{document}